# Parallel ACO with a Ring Neighborhood for Dynamic TSP

Camelia-M. Pintea[a], Gloria Cerasela Crişan[b], Mihai Manea[b]

[a]*Tech Univ Cluj Napoca, North Univ Center Baia Mare, Romania*
[b]*Faculty of Science, "Vasile Alecsandri" University of Bacău, Romania*

**Abstract**

The current paper introduces a new parallel computing technique based on ant colony optimization for a dynamic routing problem. In the dynamic traveling salesman problem the distances between cities as travel times are no longer fixed. The new technique uses a parallel model for a problem variant that allows a slight movement of nodes within their Neighborhoods. The algorithm is tested with success on several large data sets.

*Keywords:* Artificial Ants, Traveling Salesman Problem, Dynamic Problems

## 1. Introduction

*Ant Colony Optimization (ACO)* [9] is nowadays one of the metaheuristics able to solve large scale optimization problems. Ants are social insects with cooperative and adaptive features. The ant-based algorithms simulate and exploit their natural mechanisms. In order to improve the efficiency, the software designers are now using parallelization of algorithms. Ant-based algorithms are in particular well suited for parallel implementations [21] because the ants are independent and are able to operate in an asynchronous way. An overview of the parallel metaheuristics is in [5, 1] and an overview of the ACO parallel approaches is in [27].

Usually, the metaheuristics, involving intelligent [9] and complex agent-based systems [15] are successfully applied for static NP-hard problems. Nowadays the focus is on the dynamic variants of difficult optimization problems. Following this trend, a new parallel approach of the ACO for the *Dynamic Traveling Salesman Problem (DTSP)* is introduced in the current paper.

Dynamism has different meanings when referring to optimization problems. In our paper the dynamism means that from time to time a randomly chosen node is slightly moved in its neighborhood. In real life also sometimes roads are closed, or a customer is not available, forcing the supplier to deliver the goods to a nearby depot. The *DTSP* problem is about finding a minimum cost tour [5] passing exactly once through each available city [9].

The current paper starts with the state-of-art on static and dynamic ant-algorithms for *Traveling Salesman Problem (TSP)*. The new introduced parallel approach for the *Dynamic TSP* is further described and followed by numerical experiments.

*Email addresses:* cmpintea@yahoo.com (Camelia-M. Pintea), ceraselacrisan@ub.ro (Gloria Cerasela Crişan), manea_a_mihai@yahoo.com (Mihai Manea)

The paper concludes by comparatively discussing the results provided by both the sequential and parallel approaches and pointing out some further research.

## 2. THE ANT COLONY OPTIMIZATION METAHEURISTIC

*Ant colony optimization (ACO)* [6, 10] is a sub-domain of Swarm intelligence. Swarm intelligence algorithms are based on the observation of swarm's behavior, specifically on their cooperation through self organization. Many *Combinatorial Optimization Problems (COPs)* are solved today using *ACO*. In order to use *ACO* it is important to define an adequate model for a *COP*. The model of an optimization problem can be defined as in [6] and [8]:

*Definition 1*: A *model $P = (S; \Omega; f)$ of a COP* consists of:

- a search space $S$ defined over a finite set of discrete decision variables,
- a set $\Omega$ of constraints among the variables,
- an objective function $f: S \to R^+$ to be minimized.

Given a set of discrete variables $X_i$, $i = 1, ..., n$, with values $v_{ij} \in D_i = \{v_{i1}, ..., v_{i|D_i|}\}$ a *variable instantiation*, that is, the assignment of a value $v_{ij}$ to a variable $X_i$, is denoted by $X_i \leftarrow v_{ij}$.

The *search space* $S$ is the product of all sets of possible instantiations, for all the discrete variables:

$$S = (X_1 \leftarrow v_{1k_1}, X_2 \leftarrow v_{2k_2}, \ldots X_n \leftarrow v_{nk_n}),$$

$$1 \leq k_i \leq |D_i|, \forall\, 1 \leq i \leq n.$$

A *feasible solution* $s \in S$ is a complete assignment in which each decision variable has an assigned value that satisfies all the constraints in the set $\Omega$.

If the set $\Omega$ is empty, $P$ is called an unconstrained problem model, otherwise it is said to be constrained.

A feasible solution $s* \in S$ is called a *global optimum* if and only if $f(s*) \leq f(s)$, for any other feasible solution $s$. The set of all globally optimal solutions is denoted by $S^* \subseteq S$. Solving a $COP$ requires finding at least one $s* \in S^*$.

Every optimization problem can be described as a minimization problem, by appropriately changing the objective function.

The $ACO$ metaheuristic formalization follows the biological model and consists of an initialization step and a loop over three basic components: *construct ant solution, apply local search* and *update pheromone* trails. An iteration of a loop consists of constructing solutions involving all ants, the optional use of a local search algorithm and the update of the pheromones trails [8]. These procedures are detailed in the following.

- *Construct Ant Solutions.* It considers $m$ artificial ants, each constructing one solution to the problem, from elements of a finite set of available solution components $C = \{c_{ij}, i = 1, \ldots, n; j = 1, \ldots, |D_i|\}$. The construction of a solution starts with an empty partial solution $s^p = \oslash$ and at each construction step, the current partial solution $s^p$ is extended. A feasible solution component from the set of feasible neighbours $N(s^p) \subseteq C$ is repeatedly added to the current partial solution. Constructing the solutions is a process of creating a path in the graph $G_C = (V, E)$. The allowed paths in $G_C$ are defined by the solution construction mechanism that designs the set $N(s^p)$. At each step, the choosing a solution component from $N(s^p)$ is managed by probabilistic rules. The probabilistic rules are specifically defined for different $ACO$ variants. The values returned by the specific probabilistic function are using the matrix of current pheromone values ($\tau_{ij}$), but some heuristic information also.

- *Apply Local Search.* After the solutions have been constructed, many $ACO$-based heuristics make use of some optional modules. They can be used for specific actions. In most cases, at this point, a local search procedure is inserted, in order to improve the quality of the constructed solutions and usually it modifies them [8].

- *Update Pheromones.* Pheromone updates are used to decrease the pheromone values on the unused paths and also to increase the pheromone values associated with the edges of good solutions:
  - the decrease of all the pheromone values is made through *pheromone evaporation rate* $\rho \in (0\,1]$;
  - the increase of the pheromone levels is associated with the set of good solutions $S_{upd}$:
    $\tau_{ij} \leftarrow (1-\rho)\tau_{ij} + \rho \cdot \sum_{s \in S_{upd}|c_{ij} \in s} F(s)$.

where $S_{upd}$ is the set of solutions used for the update and $F : S \to R^+$ is the fitness function such that $f(s) < f(s') \Rightarrow F(s) \geq F(s')$ for all $s \neq s' \in S$.

Instantiations of this update rule are obtained by different $S_{upd}$, which is a subset of $S_{iter} \cup \{s_{bs}\}$, where $S_{iter}$ is the set of solutions constructed in the current iteration, and $s_{bs}$ is the best solution found since the beginning of the algorithm [8]. An example of update rule is the $AS$ *rule*, used in the first $ACO$ algorithm, *Ant System* ($AS$) [11], that states that

$$S_{upd} \leftarrow S_{iter}.$$

In practice it is often used an iteration best update rule (*IB*-update rule). The advantage of the *IB* rule is that is introducing a stronger bias toward the good solutions found versus the AS-update rule.

$$S_{upd} \leftarrow \arg\max_{s \in S_{iter}} \{F(s)\}.$$

The disadvantage of the *IB-update rule* is that increases the speed with which good solutions are found and also increases the probability of premature convergence. Another update rule used in practice is the best so far solution $s_{bs}$ update rule (BS-update rule). In this case, $S_{upd}$ is $\{s_{bs}\}$.

Using *IB-update rule* or *BS-update rule*, and also mechanisms for avoiding premature convergence, the ACO algorithms have better results than using the *AS-update rule*.

An efficient form of favouring the exploration of new areas in the search space, and the pheromone evaporation avoid the rapid convergence of the algorithm. For each variant of ant algorithms, such as *Ant Colony System* ($ACS$) [7, 24], or *MAX-MIN Ant System* ($MMAS$)[32], the pheromone is updated in specific way.

### 2.1. ACO for the Traveling Salesman Problem

For the purpose of this paper, we used one of the most studied $COP$: *the Traveling Salesman Problem (TSP)*. Informally, the *TSP* consists of a set of cities that had to be all visited just once and only once by a traveling salesman. The task is to find such a tour of minimal length. *TSP* is a *NP-hard* problem [18], under heavy investigation, as it easy to formalize and to represent as data structures, it has many variants due to real-world situations and it has extremely wide applications. The construction graph is fully connected and is defined by associating the set of cities with the set of vertices. The graph has the number of vertices equal to the number of cities and the lengths of the edges are proportional to the distances between the cities. When using $ACO$ to solve the $TSP$, each artificial ant incrementally constructs a solution. The ant's pheromone deposit is associated with the set of edges forming the tour it finds. The moves between cities become solution components [8].

The procedure *Ant Colony Optimization* for *Traveling Salesman Problem* ($ACO$ $TSP$) is further detailed [8].

The moves of ants between the cities define the solution components: each move from city $i$ to city $j$ incrementally adds the city $j$ to the partial solution. The length symmetry is a consequence of $c_{ij} = c_{ji}$, for any cities $i$ and $j$. The fully connected graph $G_C = (V, E)$ is defined by taking $V$ as the set of cities, and $E$ as all possible connections. The lengths of the edges between the vertices of $G_C$ are proportional to the distances between the cities represented by the vertices. The quantity of pheromone is associated with the set of edges, $E$, of the graph $G_C$.

Another perspective is the association of the set of solution components with the set of vertices instead of the set of edges in the construction graph. For the current problem, it is possible to associate the moves between nodes with the set of vertices of the graph and the locations with the set of its edges. In this way the ants' solution construction process is modified: the ants move from vertex to vertex of the construction graph choosing the connections between the nodes. In practice both perspectives are used. The current implementation uses the first perspective. The pseudocode description of $ACOTSP$ follows.

---
**Algorithm 1** Procedure ACOTSP

Initial try
**while** (terminate condition not meet) **do**
    Construct solutions
    *(each ant constructs and saves its tour)*
    Local Search *(optional)*
    Update statistics
    *(Manage some statistical information,*
    *especially if a new best solution*
    *(best-so-far or restart-best) is found*
    *and adjust some parameters*
    *if a new best solution is found)*
    Update Pheromone trail *(pheromone deposit)*
    Search control and statistics
    *(occasionally computes statistics and*
    *checks if the algorithm converged)*
**end while**

---

In ACOTSP, the solutions of the *Traveling Salesman Problem* are successively constructed as it follows:

- Every ant starts from a random node of the graph $G_C$.

- At each construction step, an ant moves along the edges of the graph.

- An ant stores in "memory" its path through the graph and chooses among the edges that do not lead to already visited vertices.

- An ant constructs a solution after it has visited all the vertices of the graph.

- At each step an ant probabilistically chooses the edge to follow among the available ones.

- After all the ants completed their tour, the pheromone on the edges is updated.

The results of $ACOTSP$ are quite good [31]. In the literature were proposed several variants of $ACO$. In the first $ACO$, *Ant System*, when constructing the solutions, the ants traverse a construction graph and make a probabilistic decision at each vertex. The pheromone values are updated by *all* the ants that have completed a graph tour.

*Max-Min Ant System* [32] is based on Ant System and introduces some important changes:

- After any iteration, only one single ant adds pheromone in order to exploit the best solution: either the one found during the current iteration (the ant which found the best solution in the current iteration), or the one that is the overall best (the ant which found the best solution from the beginning of the trial).

- The range of possible pheromone trails on each solution component is limited to an interval in order to avoid search stagnation.

- The pheromone trails are initialized with the maximum value of the pheromone interval in order to achieve higher exploration of solutions at the start of the algorithm.

The minimal and maximal pheromone values allowed are experimentally chosen based on the given problem. The maximum value could be analytically calculated based on the optimum solution length (if known). The minimum pheromone value, having a strong influence on the algorithm performance, should be based on the probability that an ant constructs the best tour found so far [32]. For some problems the choice of an appropriate minimal value is more easily experimentally done than analytically.

The process of pheromone update in *MMAS* is concluded by verifying that all pheromone values are within the imposed limits. *MAX-MIN Ant System* has significantly better results than *Ant System* for several problems, including *TSP* [32].

## 3. THE DYNAMIC TRAVELING SALESMAN PROBLEM

The definitions of dynamic traveling salesman problem and the definition of dynamic optimization problems are based on [17].

*Definition 2*: Given $n$ cities $\{c_1, c_2, ..., c_n\}$ and a cost (distance) matrix $D = \{d_{ij}\}_{n \times n}$, where $d_{ij}$ is the cost (distance) from $c_i$ to $c_j$, the *Traveling Salesman Problem (TSP)* seeks to find a permutation $\pi = (\pi_1, \pi_2, ..., \pi_n)$ that minimizes the *length of the tour corresponding to $\pi$*, denoted by:

$$\min_{\pi} \left\{ \sum_{i=1}^{n} d_{\pi_i \pi_{i+1}} \right\},$$

where $\pi_{n+1} = \pi_1$.

*Definition 3*: A *dynamic TSP (DTSP)* is a *TSP* with a dynamic cost (distance) matrix $D = \{d_{ij}(t)\}_{n(t) \times n(t)}$ where $d_{ij}(t)$ is the cost from city $c_i$ to city $c_j$, and $t$ is the real world time. The number of cities $n(t)$ and the cost matrix are time-dependent.

*Definition 4*: A dynamic optimization problem is an optimization problem with a dynamic objective function $f(t, s)$, where the goal is to minimize $f(t, s)$ for all values of $t$ (the real world time).

### 3.1. Dynamic ACO for the Traveling Salesman Problem

In [20] is proposed a hybrid technique based on *n-OPT* and *GA* for solving dynamic traveling salesman problem. *2-OPT* and *3-OPT* are used in *GA* procedures of mutation and selection. The dynamic characteristics are: the actuality (*DTSP* changes with the time), continuity (partially and quantitatively changes are done during the time), robustness (it faces unexpected situations), and effectiveness (it finds optimal tours in reasonable time).

A hybrid of the *elastic net* method and *Inver-Over* algorithm is proposed in [34] for solving *DTSP*. In this case, the heuristic information used in moving from a city to another is computed based on the standard deviation of the *Gaussian*.

An ant-based system for a dynamic *Traveling Salesman Problem*, where the travel times between the cities are changing is considered in [13]. Here the best way to handle dynamism is considered the strategy of smoothing pheromone values only in the area considered to improve the result. Other ant-based approach is the dynamic railway traveling salesman problem presented in [25]. The new information is received as time progresses and is dynamically incorporated into an evolving schedule. The ant-based procedure uses a shaking [13] technique in order to smooth all the pheromone levels. In [13] the local shake algorithm successfully combines the exploitation (using the pheromone matrix) and biased exploration within the local area.

In [14] an effective metaheuristic algorithm based on ant colony system for solving the dynamic generalized *Traveling Salesman Problem* is shown. For generalized problems, where the nodes are clustered, the dynamism is when a randomly-chosen cluster is missing from a traveling salesman tour. In this way, in the dynamic version, the distance between nodes as travel times is no longer fixed.

In the generalization of the *Vehicle Routing Problem* [23] where the nodes of a graph are partitioned into a given number of nodes sets (clusters), the objective is to find the optimal route from the given depot to the number of predefined clusters including exactly one node from each cluster. In the *dynamic Generalized Vehicle Routing Problem* [26] a variation of the *GVRP* assumes that the distances between nodes are no longer fixed. An ant-based colony heuristic was involved to solve the dynamic version of the *Generalized Vehicle Routing Problem*.

In [16] is introduced a parallelized form of the multi-algorithm co-evolution strategy for *Dynamic Multi-Objective TSP*, called synchronized parallel multi-algorithm solver. The authors aimed to show that the synchronized parallel multi-algorithm solver can be used to efficiently track the *Pareto* set, even for large instances, using multi-processor systems with shared memory.

## 4. THE NEW PARADIGM: PARALLEL ACO WITH A RING-NEIGHBORHOOD FOR DYNAMIC TSP

Serial computing has its limits, both at physical and economic level. Using a single computing system becomes inefficient when approaching the complex, highly dimensioned and large-sized problems we are facing today. The new features that parallel and distributed computing offer today open new paradigms in approaching the real-life variants of COPs.

The current paper introduces a new parallel algorithm, for solving a dynamic routing problem: the dynamic traveling salesman problem. The dynamicity is defined by changing the coordinates of a node in a specific Neighborhood.

### 4.1. Parallel ACO for the Traveling Salesman Problem

Parallel computing is an evolution of serial computing, based on what has always been the status-quo in the natural world: many complex, interrelated events happening at the same time, yet within a sequence [2]. A parallel application is designed to concurrently use the resources of a multi-core processor, or a multiprocessor computer, in an attempt to avoid performance bottlenecks. When devising a parallel application, there are supplementary challenges to address. The multiple processes that share common data have to correctly use and modify them (this is the thread-safe property). The resources can suddenly drop, or the workload can dynamically grow, so the application has to efficiently adjust (this is the scalability property). The communication between processes has to address the exchanged information (the content), the events that trigger the exchange (the timing), the pairing of exchanging processes (the connectivity), the characteristics of communications (the mode), the purpose of the received information (the exploitation), and the further processing of the information (the scope) [3].

There are many parallel approaches to TSP. From the first attempt [30], many researchers continuously investigate how different communication strategies and work balances influence the quality of the solutions and/or the speed in finding the optimum solution. The initial parallelization method used the independent island model, where several colonies solved (with different parameters) the same instance, without any communication. At the end, the best solution from all the colonies is reported as the application's result [30].

Later approaches implemented communication overheads, in order to share valuable search information. In [28] a master-slave model is implemented, with MPI communications between processes. The fact that two slave processes are involved in the communication can induce a delay of the application: one process can be faster, but it is forced to wait for the other one. As later described, our application connects the slaves only with the master, whose main task is to collect the good solutions and to control the overall behavior. In [12] a ring topology is used for communications, with good results on both *TSP* and *Vehicle Routing Problem* (*VRP*) instances. Some synchronous and asynchronous communication models that exchange the best-so-far solution are studied in [3] and a centralized control framework is proposed in [4].

Another idea for designing parallel approaches is the use of problem decomposition. One such method is to split the graph that represents the problem and to assign each such part to a processing unit. In order to solve the complete problem, an ant is allowed to move between processes, carrying its parameters and partial solution [19].

*4.2. New Parallel ACO with a Ring-Neighborhood for Dynamic TSP*

In real life, unexpected events usually trigger only local modifications. For example, a closed port due to a storm makes the cargo ship to shore and to unload the goods in the nearest port. The static, classic *TSP* is here modified in order to reflect the dynamicity of such real-life situations. Our parallel approach, based on *ACO*, is called *the Parallel ACO with a Ring-Neighborhood for Dynamic TSP (PACO–RN)*.

At the beginning of each *cycle* (made up by *Interval_Mod* iterations), one city is randomly chosen to be randomly moved in its Neighborhood. The Neighborhood is defined as the ring centred in the old position of the chosen city, having the big radius *rad* and the small radius *rad*/3 (*Figure 1*). Given a 2D *Euclidean TSP* instance, we computed $xM$, as the length of the smallest interval that contains all the abscissas of the cities, and $yM$ as the length of the smallest interval containing all the ordinates of the cities. The value *rad* is 10% of the average of $xM$ and $yM$ and is used to construct the Neighborhood of each city. The pseudocode for the parallel application follows in Algorithm 2.

The modification of the position of the city, in his specific Neighborhood, has a small impact on the tour lengths, but completely changes the theoretical notion of global optimum solution. As we continuously modify the instance, there is no optimum in the classical approach. The value delivered by our application is the less value found, but it can only be interpreted as the best length of a Hamiltonian tour that connects the cities that are "almost" fixed. The parallel application uses unevenly-balanced processes, following a master-slave model in order to solve our new *DTSP* instance. Process 0 (master) collects a pool of good solutions from the slaves and also solves the instance.

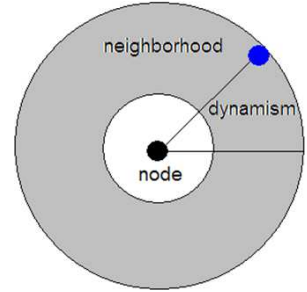

Figure 1: Neighborhood of a given node and the dynamic choice of a new node in the ring Neighborhood

---

**Algorithm 2** PACO-RN for Dynamic TSP

   Interval_Update = Interval_Mod/4
   Initial try
   **while** (termination condition not met) **do**
      **if** the *id* of the process is 0 **then**
         Slightly move one city
         Broadcast the modification
         Restore top 10 solutions
      **else**
         Modify the instance
         Compute the length of the *best so far ant*
      **end if**
      Construct solutions
      Local Search (optional)
      Update statistics
      Update Pheromone trail
      Search control and statistics
      **if** the *id* of the process is not 0 **then**
         Send the *best so far ant*
         Receive the *best top10 ant*
         *best so far ant* ← *best top10 ant*
      **else**
         Receive the *best so far ant*
         Restore top 10 solutions
         Send the *best top10 ant*
      **end if**
      Increase iteration
   **end while**

The slaves (the other processes) solve the instance and periodically send their best solutions to the master. After each good solution gathering from every slave, the master re-orders the pool of solutions and sends back to each slave the best solution he currently holds.

Our parallel application uses the sequential ACOTSP solving module described in [29] and the communications are implemented using MPICH2, a high-performance and widely portable implementation of the *Message Passing Interface (MPI)* standard (MPICH2).

## 5. NUMERICAL EXPERIMENTS

In the current section the comparisons between numerical experiments on serial and parallel implementation of *ACO* algorithm are illustrated.

The new parallel approach introduced in this paper is based on the serial *ACOTSP* software package developed in *ANSI C* by *Thomas Stützle*, freely available subject to the *GNU General Public License* [29]. The *ACOTSP* 1.0 software provides an implementation of various *Ant Colony Optimization* algorithms: *Ant System, Elitist Ant System, MAX-MIN Ant System, Rank-based* version of *Ant System, Best-Worst Ant System* and *Ant Colony System* for the symmetric *Traveling Salesman Problem (TSP)*.

The dynamic *TSP* instances used here are based on the *2D Euclidean* pcb3038, fl3795 and rl5915 from [33]. Each instance is specified by the number of nodes and the two-dimensional coordinates for each node.

The application was ran for five times using *MAX-MIN Ant System* [32] with *3-opt* local search, $\alpha=1$, $\beta=5$, $\rho=0.2$, *50* ants and *30* minutes runtime on two parallel processes. The computer used was an *Intel dual-core PC* with *2.40 GHz* clock speed and *2GB RAM*.

The averaged results and the comparison with the serial method (one process solves the dynamic *TSP* modified instance) are shown in Table 5. The first column specifies the instance and the type of application. The next three columns show the quality of the pool of best solutions: the quality of the best solution, the average quality of the best three ones, and the average quality of the complete pool, all specified as the percentage difference from the known deterministic optimum.

As Table 5 shows, the parallel application always provided better results than the serial one. The better quality is a result of cooperation of the processes, as they were able to share valuable information using the communication features provided by *MPI* commands.

Table 5 shows the average time needed to obtain the best three solutions and all the solutions from the pool. Also, Table 5 illustrates how many iterations manages the application to execute during 30 minutes runtime. The time needed to find the good solutions was less in the serial case, as all the communication commands were deleted.

| Instance | Best | Avg. 3 best | Avg. 10 best |
|---|---|---|---|
| pcb3038 | | | |
| *parallel* | **0.19** | **0.36** | **0.88** |
| *serial* | 0.34 | 0.54 | 1.29 |
| fl3795 | | | |
| *parallel* | **8.70** | **9.51** | **10.97** |
| *serial* | 9.71 | 10.43 | 12.39 |
| rl5915 | | | |
| *parallel* | **1.24** | **1.36** | **1.75** |
| *serial* | 1.53 | 1.62 | 1.99 |

Table 1: Comparison between Parallel ACO with Ring-Neighborhood (PACO-RN) for DTSP and serial application ACOTSP (in %).

| Instance | Avg. 3 best | Avg. 10 best | # of iterations |
|---|---|---|---|
| pcb3038 | | | |
| *parallel* | 1764 | 1677 | 3730 |
| *sequential* | 1757 | 1655 | 4944 |
| fl3795 | | | |
| *parallel* | 1661 | 1513 | 2640 |
| *sequential* | 1678 | 1503 | 3144 |
| rl5915 | | | |
| *parallel* | 1746 | 1596 | 840 |
| *sequential* | 1722 | 1541 | 1092 |

Table 2: Comparison of average time for providing good solutions, in seconds, and number of iterations between Parallel ACO with Ring-Neighborhood (PACO-RN) for DTSP and serial application ACOTSP.

Although the serial variant did more iterations, it quickly finds good solutions, but it did not find better results than the parallel counterpart.

In order to improve the results, we modified the parallel application, by including on slave processes one more special ant (the *copy_ant*), which stores the best ant from the pool, sent by the master process. During one cycle of *Interval_Update* iterations, this fixed, special ant is used for the global update method specified in [32].

Another idea we have also implemented was to test the *Gather/Scatter MPI* commands instead of *Send/Receive* communication commands. The results of our tests on the *rl5915* instance are illustrated in Table 5 and Table 5.

| Method | Best | Avg. 3 best | Avg. 10 best |
|---|---|---|---|
| *SR parallel* | 1.24 | 1.36 | 1.75 |
| *SRcopy_ant parallel* | **1.00** | **1.09** | **1.40** |
| *SRcopy_ant serial* | 1.40 | 1.46 | 1.90 |
| *GS parallel* | 1.63 | 1.85 | 2.33 |
| *GScopy_ant parallel* | 1.11 | 1.22 | 1.50 |
| *GScopy_ant serial* | - | - | - |

Table 3: Comparison between SR, GS methods, with or without copy_ant for rl5915 (in %).

| Instance | Avg. time 3 best | Avg. time 10 best | # of iterations |
|---|---|---|---|
| *SR parallel* | 1746 | *1596* | *840* |
| *SRcopy_ant parallel* | 1766 | *1662* | *907* |
| *SRcopy_ant serial* | 1746 | *1535* | *1068* |
| *GS parallel* | 1732 | *1555* | *770* |
| *GScopy_ant parallel* | 1722 | *1612* | *873* |
| *GScopy_ant serial* | - | - | - |

Table 4: Comparison of average time for providing good solutions, in seconds, and the number of iterations between SR, GS methods, with or without copy_ant for rl5915.

The collective commands *Gather* and *Scatter* introduce a delay, as the number of iterations is always low. The serial variant of *GSCopy_ant* is the same with the serial *SRcopy_ant*, as both delete the *MPI* commands. As a result, the quality of the solution is always weaker.

The idea of using a new ant to store the best solution sent from the pool was very good, as this small modification always brought a best result. This behavior is explained by the constant pheromone accumulation on the same very good tour; as the instance constantly modifies, more stability on edges rich in pheromone brings more exploitation of the history of the search.

## 6. CONCLUSIONS

The real-world situations exhibit a degree of uncertainty. People are trained to cope with ambiguities, and the didactic, deterministic problems need to be modified in order to reflect the current situation we are faced. This goal is achieved by the need of introducing some new dimensions for the problems, with further challenges: the classic solving methods can become inefficient, so the researchers constantly seek for new paradigms and frameworks for them.

The current paper introduces a new parallel method for solving a dynamic *TSP*. The dynamicity is introduced by randomly moving the position of a randomly chosen node, within his Neighborhood, at the beginning of each cycle. The parallel application uses a master-slave communication model and a pool of very good solutions stored by the master process. The master process also control the overall behavior, by sending 4 times by cycle to the slaves the current *best-of-all* solution. We intend to continue the research by studying the behavior of our parallel application on large instances, exploring other dynamic variants (blocking some edges, or modifying some edge lengths), and executing the application on multiple quad-core processors, in order to better tailor the communication design.